%% file: main.tex
\documentclass[letterpaper, 10 pt, conference]{ieeeconf}  
                                                          
\usepackage{graphicx}
\usepackage{amsmath}
\usepackage{amssymb}
\usepackage{csquotes}
\usepackage{url}
\usepackage{hyperref}
\usepackage{subcaption}
\usepackage{color, colortbl}
\usepackage{algorithm}
\usepackage{algpseudocode} 
\usepackage[nolist]{acronym} 
\let\labelindent\relax
\usepackage[inline]{enumitem} 
\usepackage{cite}
\graphicspath{ {imgs/} }
\usepackage{siunitx}
\newif\ifshowComments
\showCommentsfalse

\ifshowComments
    \usepackage[draft,markup=bfit, deletedmarkup=sout, authormarkup=brackets]{changes}
\else
    \usepackage[final,markup=bfit, deletedmarkup=sout, authormarkup=brackets]{changes}
\fi

\definechangesauthor[name={Lukas},color=green]{LR}
\definechangesauthor[name={Matej}, color=orange]{MH}

\ifshowComments

\else

\fi

\input{acronyms.tex}

\input{macro.tex}

\useVspacetrue
\usepackage[firstpageonly=true]{draftwatermark}

\SetWatermarkAngle{0}
\SetWatermarkColor{black}
\SetWatermarkLightness{0.5}
\SetWatermarkFontSize{9pt}
\SetWatermarkVerCenter{30pt}
\SetWatermarkText{\parbox{30cm}{%
\centering This is the authors' final version of the manuscript published as:\\
\centering Rustler, L.; Matas, J. \& Hoffmann, M. (2023),\\
\centering Efficient Visuo-Haptic Object Shape Completion for Robot Manipulation \\
\centering  in '2023 IEEE/RSJ International Conference on Intelligent Robots and Systems (IROS)', pp. 3121-3128. (C) IEEE \\
\centering \url{https://doi.org/10.1109/IROS55552.2023.10342200}
}}

\title{Efficient Visuo-Haptic Object Shape Completion for Robot Manipulation}

\author{Lukas Rustler, Jiri Matas, and Matej Hoffmann %
\thanks{This work was supported by the OP VVV MEYS funded project CZ.02.1.01/0.0/0.0/16\_019/0000765 ``Research Center for Informatics''. L.R. was additionally supported by the Czech Technical University in Prague, grant no. SGS22/111/OHK3/2T/13.}
\thanks{Lukas Rustler, Jiri Matas, and Matej Hoffmann are with the Department of Cybernetics, Faculty of Electrical Engineering, Czech Technical University in Prague,
 {\tt\small matej.hoffmann@fel.cvut.cz, matas@fel.cvut.cz, lukas.rustler@fel.cvut.cz.}}}
 
\IEEEoverridecommandlockouts
\begin{document}

\maketitle

\begin{abstract}
For robot manipulation, a complete and accurate object shape is desirable. Here, we present a method that combines visual and haptic reconstruction in a closed-loop pipeline. From an initial viewpoint, the object shape is reconstructed using an implicit surface deep neural network. The location with highest uncertainty is selected for haptic exploration, the object is touched, the new information from touch and a new point cloud from the camera are added, object position is re-estimated and the cycle is repeated. We extend \etal{Rustler} (2022) by using a new theoretically grounded method to determine the points with highest uncertainty, and we increase the yield of every haptic exploration by adding not only the contact points to the point cloud but also incorporating the empty space established through the robot movement to the object. Additionally, the solution is compact in that the jaws of a closed two-finger gripper are directly used for exploration. The object position is re-estimated after every robot action and multiple objects can be present simultaneously on the table. 
We achieve a steady improvement with every touch using three different metrics and demonstrate the utility of the better shape reconstruction in grasping experiments on the real robot. On average, grasp success rate increases from 63.3\% to 70.4\% after a single exploratory touch and to 82.7\% after five touches. The collected data and code are publicly available (\url{https://osf.io/j6rkd/}, \url{https://github.com/ctu-vras/vishac}).


\end{abstract}

\input{Sections/introduction}

\input{Sections/related_work}
\input{Sections/method}
\input{Sections/experiments}
\input{Sections/conclusion}

\bibliographystyle{IEEEtran}
\bibliography{refs}

\end{document}

%% file: acronyms.tex
\newacro{dnn}[DNN]{Deep Neural Network}
\newacro{fcn}[FCN]{Fully Convolutional Network}
\newacro{pc}[PC]{Point Cloud}
\newacro{sdf}[SDF]{Signed Distance Function}
\newacro{cnn}[CNN]{Convolutional Neural Network}
\newacro{gnn}[GNN]{Graph Neural Network}
\newacro{dl}[DL]{Deep Learning}
\newacro{ml}[ML]{Machine Learning}
\newacro{gpis}[GPIS]{Gaussian Process Implicit Surface}
\newacro{mc}[MC]{Monte Carlo}
\newacro{mlp}[MLP]{Multi-Layer Perceptron}
\newacro{bpa}[BPA]{Ball Pivoting Algorithm}
\acrodefplural{gpis}[GPIS's]{Gaussian Process Implicit Surfaces}
\newacro{gpisp}[GPISP]{Gaussian Process Implicit Shape Potential}
\newacro{gp}[GP]{Gaussian Process}
\acrodefplural{gp}[GPs]{Gaussian Processes}
\newacro{ros}[ROS]{Robot Operating System}
\newacro{icp}[ICP]{Iterative Closest Point}
\newacro{fps}[FPS]{Farthest Point Sampling}
\newacro{igr}[IGR]{Implicit Geometric Regularization for Learning Shapes}
\newacro{js}[JS]{Jaccard similarity}
\newacro{cd}[CD]{Chamfer distance}

\newacro{nerf}[NeRF]{Neural Radiance Fields}

%% file: macro.tex
\newcommand{\equationref}[1]{\hyperref[#1]{Eq.~\ref*{#1}}}
\newcommand{\figref}[1]{\hyperref[#1]{Fig.~\ref*{#1}}}
\newcommand{\tabref}[1]{\hyperref[#1]{Table~\ref*{#1}}}
\newcommand{\secref}[1]{\hyperref[#1]{Section~\ref*{#1}}}
\newcommand{\algoref}[1]{\hyperref[#1]{Alg.~\ref*{#1}}}

\newcommand{\norm}[1]{\left\lVert#1\right\rVert}

\DeclareMathOperator{\E}{\mathbb{E}}
\newcommand{\abs}[1]{\left\lvert#1\right\rvert}

\newcommand{\argmax}{\operatornamewithlimits{argmax}}

\newcommand{\etal}[1]{#1 \textit{et al.}}

\newif\ifuseVspace
\newcommand{\customVspace}[1]{\ifuseVspace \vspace{#1} \fi}

\def\methodname{VISHAC}

\def\ie{, \textit{i.e.}, }
\def\eg{, \textit{e.g.}, }

\def\pc{point cloud}
\def\pcs{point clouds}

\def\gt{ground truth}

%% file: Sections/introduction.tex
 \section{Introduction}
We consider the following robotic setup. A static RGB-D camera connected to a robotic arm controller is observing one or more unknown 3D objects. 
To be able to grasp and manipulate the object, the robotic system needs a model of the object in terms of a complete shape\eg{}an accurate mesh. 
There are intrinsic limitations to the performance of computer vision techniques for 3D reconstruction of objects from images or point clouds if only a limited number of viewpoints is available. Solutions relying on RGB or RGB-D images, LIDAR point clouds, or voxels, cannot easily overcome self-occlusion and may have specific difficulties with transparent or specular objects. 
The robot arm cannot reliably grasp and manipulate the object given only partial information. However, the manipulator can be controlled to touch or poke the object in order to extend the surface for which the model is accurate. 

We address the following problem. Given an initial \mbox{RGB-D} map, obtain an accurate representation of the complete shape of the object with the help of exploratory contact actions. The objective is either to maximize the accuracy given an upper bound on the number of touches or minimize the number of touches to reach a predefined accuracy on the complete surface. This problem in turn requires solutions of sub-problems such as prediction of the least reliable part of the surface, estimation of the free space around the objects and the detection of touch-induced object movement.

\begin{figure}[t!]
    \centering
    \includegraphics[width=0.49\textwidth]{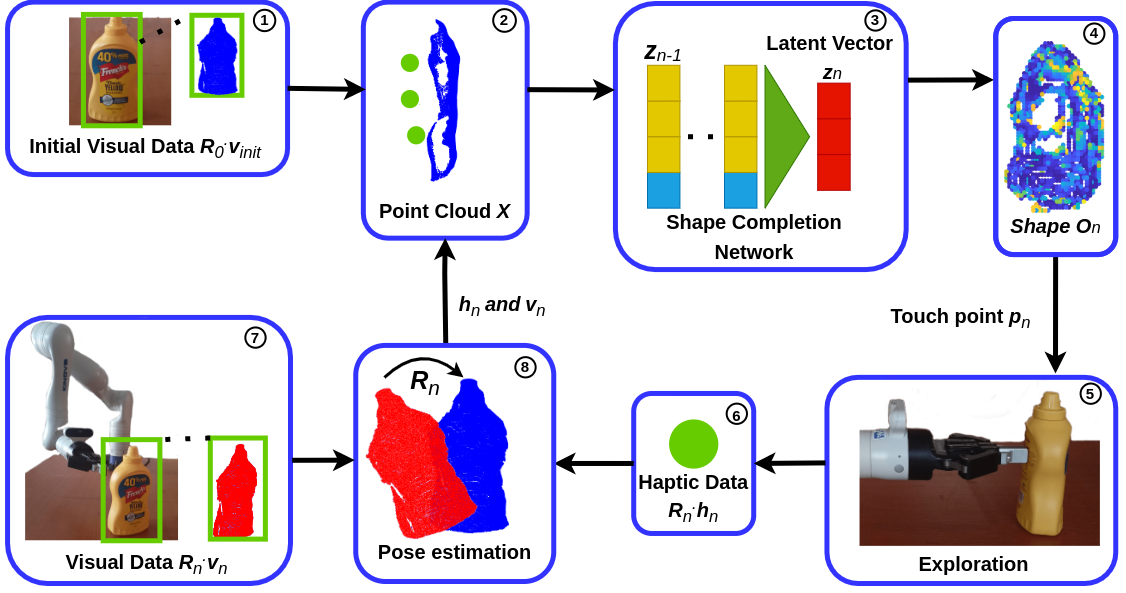}
    \caption{Schematic operation of \methodname{}. An initial \mbox{RGB-D} image of the scene is captured (1), a transformation $\mathbf{R}_0$ from the robot base to the object is obtained, and the object is segmented and converted into a \pc{} $\mathcal{X}$ (2). Iterative reconstruction: In each step, $n=0:(N-1)$, the \pc{} is inserted into a neural network (3) and a completed shape $\mathbf{O}_n$ is created (4). The most uncertain point $\mathbf{p}_n$ is selected for touch (5). After contact, the object may have been displaced, giving rise to a new transformation $\mathbf{R}_n$. Haptic data $\mathbf{h}_n$ (6) from contact and visual data by taking a new image from the RGB-D camera $\mathbf{v}$ (7) are collected. The transformation $\mathbf{R}_n$ is computed from pose estimation (8) and the new data, transformed into the original frame $\mathbf{R}_0$, are added to $\mathcal{X}$. See Sec.~\ref{sec:algo} for details.}
    \label{fig:schema}
    \customVspace{-2em}
\end{figure}

The scenario models a setup, where new objects are presented to a system that minimizes the risk of an unstable grasp, and therefore ``explores'' the shape of the object by touching and poking before attempting a grasp and subsequent manipulation. For example, imagine a conveyor belt for sorting objects of different sizes into respective bins, e.g. in a scrapyard, where the robot must be able to pick up any object.   

\textbf{Contributions.} We present a pipeline for visuo-haptic shape completion called \methodname{}\replaced{. We extend the baseline by \etal{Rustler}~\cite{Rustler2022}, which also deals with shape completion. We describe our modifications and improvements with respect to \cite{Rustler2022} below.}{, importantly extending the work of \etal{Rustler}~\cite{Rustler2022}, which serves as a baseline here.}

The first group of improvements concerns the process of shape completion performed by \ac{igr}, which we modified as follows. We use a new, theoretically grounded, method to determine the points with highest uncertainty. In addition, we changed the sampling of points inside the network to respect the input \pc{} more closely. Finally, the yield of every haptic exploration is increased by adding not only the contact points to the \pc{} but also incorporating the empty space established through the robot movement to the object. The two last mentioned improvements together make the pipeline more robust.
The second group of enhancements pertains to the practical aspects of this scenario. We do not use a dedicated ``finger'' for haptic exploration but directly the jaws of a closed two-finger gripper, which is a more compact solution. The object position is newly re-estimated after every robot action, so objects are allowed to move after being poked. Multiple objects can be present simultaneously on the table. The speed of the pipeline is improved through parallelization (2 times in simulation and 2.4 times in the real world).  We contribute a more detailed evaluation using a set of different metrics. Finally, real-word grasping experiments demonstrate the effectiveness of our approach.

The collected data and code are publicly available {\footnotesize{(\url{https://osf.io/j6rkd/}, \url{https://github.com/ctu-vras/vishac})}}. An accompanying video can be found at {\footnotesize{\url{https://youtu.be/V5mNKDACetA}}}.


%% file: Sections/related_work.tex
\section{Related work}
For the problem addressed, two main sources of information have been used: visual input (RGB or RGB-D sensors) and haptic exploration (tactile or force sensing). In recent work, the two have often been combined. 

\subsection{Visual-only Shape Completion}
Devices able to sense depth and thus create \pcs{} are widely available and can provide rich information about the scene. The early methods were based on geometric properties or templates. The geometric ones either assume that most objects humans use are symmetric, so completion can be done by mirroring partial information about its axis of symmetry~\cite{bohg2011mind}, or detect primitive shapes~\cite{schnabel2009completion}. Template-based methods benefit from prior knowledge in the form of a database of object shapes~\cite{pauly2005example}. 

More recently, solutions based on machine learning gained popularity. An example is \acl{gpis}~\cite{li2016dexterous}, which, however requires points on the whole surface and suffers from poor scaling over a dense \pc{}. Thus, the input must be downsampled, resulting in loss of detail. Other methods were originally created to make surfaces from complete \pcs{}, but provide shape completion abilities by interpolating between shapes in a latent space~\cite{parkDeepSDFLearningContinuous2019,atzmonSALSignAgnostic2020,groppImplicitGeometricRegularization2020}. In this work, we start from \ac{igr}~\cite{groppImplicitGeometricRegularization2020} and propose improvements. 

\acl{dl} techniques such as \acp{cnn} for shape completion typically represent objects as voxel grids. This allows to introduce probabilistic uncertainty in voxel grids, but the methods usually suffer from cubically growing computational requirements with the number of voxels, limiting the resolution of the output shape. The newest methods utilize graph attention networks~\cite{huang2021gascn} or transformers to complete a shape~\cite{Rosasco2022, Vall2022}.

\subsection{Haptic-Only Shape Completion}
With advances in haptic exploration, some purely haptic approaches have been proposed. They utilize some techniques mentioned above, such as implicit shape potentials~\cite{ottenhausLocalImplicitSurface2016}, or \acp{gp}~\cite{yiActiveTactileObject2016, driessActiveLearningQuery2017, dragiev2013uncertainty}. Gaussian-based methods have the advantage of having the ability to express uncertainty directly from their nature using the variance of each point. However, haptic-only completion needs a high number of touches, which is time-demanding.

\subsection{Visuo-Haptic Shape Completion}
A combination of visual and haptic data has the potential to combine the best of both worlds. \ac{gp} methods were proposed~\cite{gandlerObjectShapeEstimation2020,ottenhausVisuoHapticGraspingUnknown2019, bjorkmanEnhancingVisualPerception2013}. However, points covering most of the surface are needed, which leads to the need for a lot of exploration. A way to overcome this issue may be to use symmetry as in \cite{Bonzini2022}.

\ac{cnn}-based methods~\cite{watkins-vallsMultiModalGeometricLearning2019,wang3DShapePerception2018a} usually require fewer touches but suffer from lower resolution due to computational requirements. \etal{Smith} proposed approaches \cite{smith3DShapeReconstruction2020,smithActive3DShape2021} based on \acl{gnn}. Reconstructions by these methods have a higher resolution but are nonsmooth and, for now, evaluated only in simulation.

An important part of haptic exploration is the decision where to touch. The object can be touched randomly as done by \etal{Smith}~\cite{smith3DShapeReconstruction2020}, or always select a position opposite the camera (from \enquote{behind}) as \etal{Watkins-Vall}~\cite{watkins-vallsMultiModalGeometricLearning2019}. However, these are not as effective as an uncertainty-driven approach. Uncertainty can come from the Gaussian distribution~\cite{bjorkmanEnhancingVisualPerception2013, wang3DShapePerception2018a, gandlerObjectShapeEstimation2020, ottenhausVisuoHapticGraspingUnknown2019, Bonzini2022}; from the Monte Carlo dropout~\cite{Murali2022}; \added{from \ac{nerf}~\cite{pan_activenerf, jin_neunbv};} or from the \ac{sdf}~\cite{Rustler2022, Suresh2022}. Alternatively, it can be learned where to touch as in \etal{Smith}~\cite{smithActive3DShape2021}.

Our work belongs to uncertainty-driven approaches and is based on implicit surface deep neural network \ac{igr}~\cite{groppImplicitGeometricRegularization2020}, exploiting the definition of \ac{sdf} to estimate uncertainty in order to efficiently explore the most promising parts of the objects. We extend and directly compare ourselves with \cite{Rustler2022}. Indirectly, this encompasses also a comparison with other methods, namely \cite{bernardiniBallpivotingAlgorithmSurface1999, kazhdan2006poisson, guoSurfaceReconstructionUsing1997, Williams2006}, which were outperformed in \cite{Rustler2022}.  

%% file: Sections/method.tex
\section{Method}
We propose an iterative method depicted in \figref{fig:schema}.
The objective is to iteratively improve shape reconstruction of objects on the table combining images and selecting locations for tactile exploration.
The algorithm is described in detail in~\secref{sec:algo}. The following sections detail individual modules required by the pipeline. 

\subsection{Implicit Surfaces}
\label{sec:implicit_surfaces}
An implicit surface is a set of points whose signed distance to a surface is equal to zero. The function to compute this distance is called \acf{sdf} and is defined as
\begin{equation}
    f(\mathbf{x}) = s,
\end{equation}
where, in our case, $\mathbf{x}$ is a point defined in 3D space and $s$ is the signed distance. Traditionally, $f$ would be described analytically, but it can also be learned with a neural network. Then, the implicit surface generated with a neural network can be described as
\begin{equation}
    \label{eq:surface}
    \mathcal{M} = \{ \mathbf{x} \in \mathbb{R}^3 \; | \; f(\mathbf{x};\boldsymbol{\theta}) = 0\}, 
\end{equation}
where $f(\mathbf{x};\boldsymbol{\theta}):\mathbb{R}^3 \rightarrow \mathbb{R}$ is a \ac{mlp} learned to approximate \ac{sdf}, with $\boldsymbol{\theta}$ being the parameters of the network.

\subsection{Implicit Geometric Regularization for Learning Shapes}
\replaced{W}{As in \cite{Rustler2022}, w}e selected \ac{igr}~\cite{groppImplicitGeometricRegularization2020} to represent the function \textit{f}. The method assumes at input a \pc{} $\mathcal{X} = \{\textbf{x}_{1:C}\}$, where $C$ is the number of points in the \pc{}, and optionally a set of normals for each point $N = \{\mathbf{n}_{1:C}\}$. To train on multiple objects, the network utilizes an auto-decoder architecture from \etal{Park}~\cite{parkDeepSDFLearningContinuous2019} with different latent code $\mathbf{z}_i$ for every shape $i\in I$ in the input set. In the prediction phase $|I| = 1$. Therefore we will introduce loss only for one shape $i\in I$, defined as:
\begin{equation}
\label{eq:rec_loss}
    \footnotesize{
    \ell(\boldsymbol{\theta}, \mathbf{z}_i)=\ell_\mathcal{X}(\boldsymbol{\theta}, \mathbf{z}_i)+ \lambda \E_\mathbf{x}{\left[\norm{\nabla_\mathbf{x}f(\mathbf{x};\boldsymbol{\theta}, \mathbf{z}_i)}-1\right]}^2 + \alpha \norm{\mathbf{z}_i}}
\end{equation}
where  
\begin{equation}
\label{eq:X_loss}
\footnotesize{
\ell_\mathcal{X}(\boldsymbol{\theta}, \mathbf{z}_i)=\frac{1}{C}\sum_{c=1}^C (\abs{f(\mathbf{x}_c;\boldsymbol{\theta}, \mathbf{z}_i)} + \norm{\nabla_\mathbf{x}f(\mathbf{x}_c;\boldsymbol{\theta}, \mathbf{z}_i)-\mathbf{n}_c})}
\end{equation}
The first term in \equationref{eq:rec_loss} encourages $f$ to vanish on $\mathcal{X}$ and $\nabla_{\mathbf{x}}f$ to be close to the supplied normals. The second term is called the Eikonal term and regularizes the network by pushing $\nabla_{\mathbf{x}}f$ to be of unit Euclidean norm. \deleted{The term is also used for uncertainty estimation---described later in \secref{sec:uncertainty}.}

The result is iteratively optimized at both train and inference time (over multiple shapes in batches from the whole train set $I$ while training and over one shape for prediction). The parameters $\boldsymbol{\theta}$ are fixed during inference and only $\mathbf{z}_i$ is changed. In this work, we use a trained network from \cite{Rustler2022} and our modifications pertain to inference only. \added{See \cite{Rustler2022} for more information about training and the architecture.}

\subsection{IGR Modifications -- Sampling and Free Space}
\label{sec:igr_changes}
By default, the \ac{igr} network uses a random sampling of points in each of its iterations. It is effective when a complete \pc{} is inserted. However, when a partial \pc{} is used, the network tends to inflate the objects\ie{}the output shape is over the boundaries of the input. We propose to use \ac{fps} as in~\cite{qi2017PointNet}. The points sampled by this algorithm are spatially far from each other and help the network better understand the whole object in each iteration, making the final shape tighter with the input. 

Another improvement of the method is the use of information about the space explored. As the robot moves through space, we know that the traversed space is free and no shape can be there. During the inference phase of the network, a signed distance is calculated for every point in the free space explored. We keep in memory only the points that are less than \SI{20}{cm} from the center of a given object. We want all free space points to be outside of the surface\ie{}to have a positive signed distance. Therefore, we add to the loss the sum of the absolute values of all distances that are lower than \SI{1}{mm}.

\subsection{Object Representation from Visual and Haptic Data}
 There are several possible representations of an object $O$. We choose to represent an object as a \pc{} (which is, in fact, an implicit surface from \equationref{eq:surface}), concatenated with uncertainty for each point. Mathematically expressed as
\begin{equation}
    \label{eq:object}
    O = \{ \mathbf{x} \in \mathbb{R}^3, u \in \mathbb{R} \; | \; f(\mathbf{x};\boldsymbol{\theta}, \mathbf{z}) = 0, u \geq 0\},
\end{equation}
where $u$ stands for the uncertainty of the given point. 

Our method is iterative, therefore, we compute a new shape in each iteration $n$. For $n=0$ the shape $O_0$ is computed only from RGB-D information $\mathbf{v}_{init}$. In all other iterations we get the shape $O_n$ with visual information $\mathbf{v}_{init}$, $\mathbf{v}_{0:(n-1)}$, and haptic information $\mathbf{h}_{0:(n-1)}$. \equationref{eq:object} is therefore changed to
\begin{equation}
    \label{eq:object_t}
    O_n = \{ \mathbf{x} \in \mathbb{R}^3, u \in \mathbb{R} \; | \; f(\mathbf{x};\boldsymbol{\theta}, \mathbf{z}_n) = 0, u \geq 0\},
\end{equation}
where $\mathbf{z}_n$ is the current latent vector optimized on \pc{} $\mathcal{X}_n = \{\mathbf{h}_0,\dots,\mathbf{h}_{(n-1)}, \mathbf{v}_0,\dots,\mathbf{v}_{(n-1)}, \mathbf{v}_{init}\}$.

To obtain haptic information $\mathbf{h}_n$, we must first select the position for haptic exploration $\mathbf{p}_n$ on the object that minimizes the global uncertainty of the object\ie{}select the point with the highest current uncertainty as
\begin{equation}
\label{eq:pn}
\begin{split}
     &m = \argmax_u {O_n}, \\
    &\mathbf{p}_n = \mathbf{x}_m \in O_n.
\end{split}
\end{equation}
The desired $\mathbf{h}_n$ is then obtained from the position of the actual contact between the robot and the object. 

Note that we showed the equations for only one object at a time. However, our method is capable of handling multiple objects in a complex scene. So, we will later in an algorithm refer to objects as $O_{k,n}$, where $k=1:K$ is the object's id in the scene, and $n$ is the current iteration.

\subsection{Object Shape Uncertainty}
\label{sec:uncertainty}
A crucial part of this work is the estimation of uncertainty. We selected part of the loss from \equationref{eq:rec_loss}, particularly the Eikonal loss
\begin{equation}
    \label{eq:eikonal}
    \ell_{Eikonal} = (\norm{\nabla_\mathbf{x}f(\mathbf{x};\boldsymbol{\theta}, \mathbf{z})}-1)^2.
\end{equation}
It was proven by Takashi~\cite{takashi1996} that a given function $f(\mathbf{x})$ that meets the condition of the Eikonal Equation $\norm{\nabla_\mathbf{x} f(\mathbf{x})} = 1$ on a Riemann manifold $M$ is a \ac{sdf} to a hypersurface $M$. Furthermore, Crandal and Lions~\cite{Crandall1983} presented Viscosity Solutions that prove the same applies even if $\norm{\nabla_\mathbf{x} f(\mathbf{x})}$ does not exist on every $\mathbf{x}$. 

\replaced{Points with a high loss from \equationref{eq:eikonal} are not on the estimated surface. And, intuitively, if all points were certain, the loss would be zero for all of them. However, this was not the case in our experiments. Therefore, we compute the Eikonal loss from \equationref{eq:eikonal} for all points on our current shape $O$ and take it as uncertainty---where the higher the loss, the higher the uncertainty. An empirical comparison of the overall uncertainty of the shapes and reconstruction quality can be seen in \figref{fig:complex}.}
{Given these results, we can compute the Eikonal loss from \eqref{eq:eikonal} for all points on our current shape $O$—where the higher the loss, the higher the uncertainty.}
\subsection{Segmentation of Multiple Objects}
\label{sec:segmentation}
First, bounding boxes of all objects in the input RGB image are found---in the real world using Yolov7~\cite{wang2022yolov7} fine-tuned on our objects, and using color-based segmentation in the simulation. We then run the Flood Fill algorithm~\cite{tint} on the depth image (aligned to RGB). The algorithm starts at a given pixel (we use the center of the bounding box) and expands over neighbors that fulfill the given criterion (difference in depth in our case) until no neighbor is left. We also restrict the region of interest by the bounding box from the RGB image. Segmented depths, together with camera information, are then used to create \pcs{} of objects. A more detailed description can be found on GitHub {\footnotesize{\url{https://github.com/ctu-vras/vishac}}}.


\subsection{Pose Estimation}
\label{sec:pose_estimation}
In haptic exploration methods, a common but unrealistic assumption is that objects are fixed to the surface (also in \cite{Rustler2022}). Objects naturally move when they are touched and their pose needs to be re-estimated. Many existing pose estimation methods require prior knowledge of the objects at the instance level \cite{PoseCNN, wang2019densefusion} or category level \cite{wang2019normalized, chen2020category}. We seek methods that work with unknown arbitrary objects. Having segmented \pcs{} of each object at hand, we chose a simple and computationally cheap (no GPU) solution using \ac{icp}~\cite{besl1992}. Alternative solutions for unknown objects are \cite{Wen2021,Li2020}.

\subsection{\methodname{} Algorithm}
\label{sec:algo}
We present the algorithm of our method in \algoref{alg:uncertainty-driven_shape completion} and the same is depicted in \figref{fig:schema}. The algorithm is high-level pseudocode, with a module for shape completion \algoref{alg:complete_shape} described in more detail.

\input{algos/shape_completion}

We will first describe the module for the shape creation itself. In~\cite{Rustler2022} the \ac{igr} network was used as a standalone library. To perform more efficiently and to be able to handle more objects at once, we modified it to be more compatible with the whole ecosystem (under \ac{ros}). The module contains the input \pcs{}, latent vectors, and other parameters for each object in the scene, allowing simple switching between objects without excessive overhead. The next object to be completed is selected through messages sent from the main script. If a new request is received and reconstruction is running, the new objects are placed in a queue. The module runs in the background, which allowed a considerable speed-up of the whole process, as now reconstructions are processed while the robot is moving. The basic operation is shown in \algoref{alg:complete_shape}. First, a new shape is selected from a queue (if it is not empty, otherwise the module waits for a new request). Then the latent vector $\mathbf{z}$ and the input \pcs{} $\mathcal{X}$ for the given shape are loaded. If the shape is new, the first latent code is created randomly with a normal distribution. Otherwise, the last known vector for the given object is used. The current $\mathbf{z}$ is optimized with the loss from \equationref{eq:rec_loss}. Finally, the shape $O$ is created, together with the uncertainty computed with \equationref{eq:eikonal}.

\input{algos/pipeline}

The main \algoref{alg:uncertainty-driven_shape completion} starts with capturing the initial visual information (box (1) in \figref{fig:schema}, line \ref{line:visu_init} in \algoref{alg:uncertainty-driven_shape completion}). An initial transformation $\mathbf{R}_0$ of the object in the base frame of the robot is obtained. The information is then segmented and a \pc{} is created for each object in the scene (box (2), line \ref{line:segment}). The segmentation itself is described in \ref{sec:segmentation}.

Every iteration starts with computation of the current pose for all objects in the scene (\secref{sec:pose_estimation}). The pose for all objects is computed here---the explored object may change pose after the touch is released and surrounding objects may have been moved unintentionally, so it is more robust to compute pose for all objects. This pose is used mainly to correctly select the point to explore.

Having the segmented \pcs{} and poses, a request for shape is sent (box (3), line \ref{line:complete_shape}). In the first iteration, we request shapes for all objects to create collision shapes for the motion planning algorithm. In other iterations, we request shape only for the last touched object. The impact position $\mathbf{p}_n$ is selected (box (4), line \ref{line:select_pn}) based on \equationref{eq:pn}. When more than one object is available, the impact position is selected as the point with maximal uncertainty among all the objects. \deleted{To prevent exploration of only one object, we allow one object to be touched three times in a row, and then it is removed from the touch-selecting algorithm for the given iteration.}

After $\mathbf{p}_n$ is selected, the robot is moved to the position and contact information is extracted (box (5-6), lines \ref{line:explo_start}-\ref{line:explo_end}). The movement consists of two subsequent movements. Firstly, the robot is moved to a position \SI{10}{cm} from object along the normal of $\mathbf{p}_n$. Next, 
the robot is moved along the normal with linear movement until contact occurs. In our case, the contact is detected with the change in joint torques. Haptic information $\mathbf{h}_n$ is created as a circle perpendicular to the impact normal, with the center in the position of the end effector.

After collision, new visual information is saved, segmented and added to the \pc{} for the touched object, together with the haptic information (box (7), lines \ref{line:after_touch_start}-\ref{line:after_touch_end}). To make sure that we segment the correct object, the RGB-D information is cropped with the bounding box found for the given object in the last iteration (the box is slightly enlarged to allow movement of the object). Finally, the pose $\mathbf{R}_n$ of the object right after touch (before the contact is released) is computed (box (8), line \ref{line:pose_est}). This pose is used to transform the current data into the frame of $\mathbf{v}_{init}$---the first frame must be used so that the latent vectors for the given objects can be reused. Note that now only the pose of the explored object is computed, unlike at the beginning of each iteration (line \ref{line:pose_start}). Also, in \figref{fig:schema}, only one pose estimation is shown for simplicity.

The whole pipeline runs until the selected number of touches (over all objects) is done or until the time limit is reached.  


%% file: algos/shape_completion.tex
\begin{figure}[tb]
\begin{algorithm}[H]
\caption{Create Shape}
\label{alg:complete_shape}
    \begin{algorithmic}[1] 
    \While {Pipeline is running}
        \State $k=\textit{SelectFromQueue()} \; or \; \textit{WaitForRequest()}$;
        \State $\mathbf{z} = \textit{LoadLatent(k)}$;
        \State $\mathcal{X} = \textit{LoadData(k)}$;
        \State $\mathbf{z}_{optimized} =  \textit{Optimize}(\mathbf{z}, \mathcal{X})$; \Comment{Loss from \equationref{eq:rec_loss}}
        \State $O=\textit{PredictShapeAndUncertainty}(\mathbf{z}_{optimized})$;
    \EndWhile
    \end{algorithmic}
\end{algorithm}
\customVspace{-2.5em}
\end{figure}

%% file: algos/pipeline.tex
\begin{figure}[tb]
\begin{algorithm}[H]
\caption{Multi-Object Shape Completion}
\label{alg:uncertainty-driven_shape completion}
    \begin{algorithmic}[1] 
    \Statex \textbf{Input:} maximal number of haptic explorations N, maximal run time $T$
    \Statex \textbf{Output:} Final shape $O_{1:K, n}$ \Comment{For $K$ shapes}
    \State $t_{init} =  \textit{CurrentTime}()$;
    \State $\text{Start } \textit{CompleteShape}()\text{ service in background}$;
        \label{line:service_start}
    \State $\mathbf{v}_{init} =  \textit{CaptureVisualInformation}()$;
        \label{line:visu_init}
    \State $\mathcal{X}_{1:K} =  \textit{Segment}(\mathbf{v}_{init})$;
        \label{line:segment}
    \State $k = 1:K$; \Comment{Start with all objects}
    \For{$n= 0,~\dots,~(N-1)$}
        \If {$ \textit{CurrentTime}()-t_{init} \ge T$}
            \State \textbf{break;}
        \EndIf
        \State $\mathbf{R}_{1:K, n}= \textit{ComputePose}(1,\dots,K)$;
            \label{line:pose_start}
        \State $O_{k,n} =  \textit{CompleteShapeRequest}(k)$; 
            \label{line:complete_shape}
        \State $\mathbf{p}_n; k= \textit{SelectTouchPoint}(O_{1:K,n}, \mathbf{R}_{1:K, n});$
            \label{line:select_pn}
        \State $ \textit{MoveRobot}(\mathbf{p}_n);$
            \label{line:explo_start}
        \State $\mathbf{R}_{k, n}\cdot \mathbf{h}_n =  \textit{GetContactInformation}();$
            \label{line:explo_end}
        \State $\mathbf{R}_{k, n}\cdot \mathbf{v}_{n} =  \textit{CaptureVisualInformation}()$;
            \label{line:after_touch_start}
        \State $\mathbf{R}_{k, n} =  \textit{ComputePose}(k)$;
            \label{line:pose_est}
        \State $\mathbf{v}_n; \mathbf{h}_n =  \textit{Transform}(\mathbf{R}_{k, n}\cdot \mathbf{v}_{n}, \mathbf{R}_{k, n}\cdot \mathbf{h}_{n}, \mathbf{R}_{k, n});$
        \State $\mathcal{X}_k \mathrel{{+}{=}}  \textit{Segment}(\mathbf{v}_n)$;
        \State $\mathcal{X}_k \mathrel{{+}{=}}  \mathbf{h}_n;$
            \label{line:after_touch_end}

    \EndFor
    \State \textbf{Return: }$O_{1:K, n}$
    \end{algorithmic}
\end{algorithm}
\customVspace{-2.5em}
\end{figure}

%% file: Sections/experiments.tex
\section{Experiments and Results}
The primary experiments we conducted demonstrate how the completeness and accuracy of reconstructions change with each additional touch.
In addition, we evaluated the quality of the reconstructions in a real-world grasping experiment. Examples are shown in the accompanying video ({\footnotesize{\url{https://youtu.be/V5mNKDACetA}}}).

\begin{figure}[tb]
    \centering
    \includegraphics[width=0.3\textwidth]{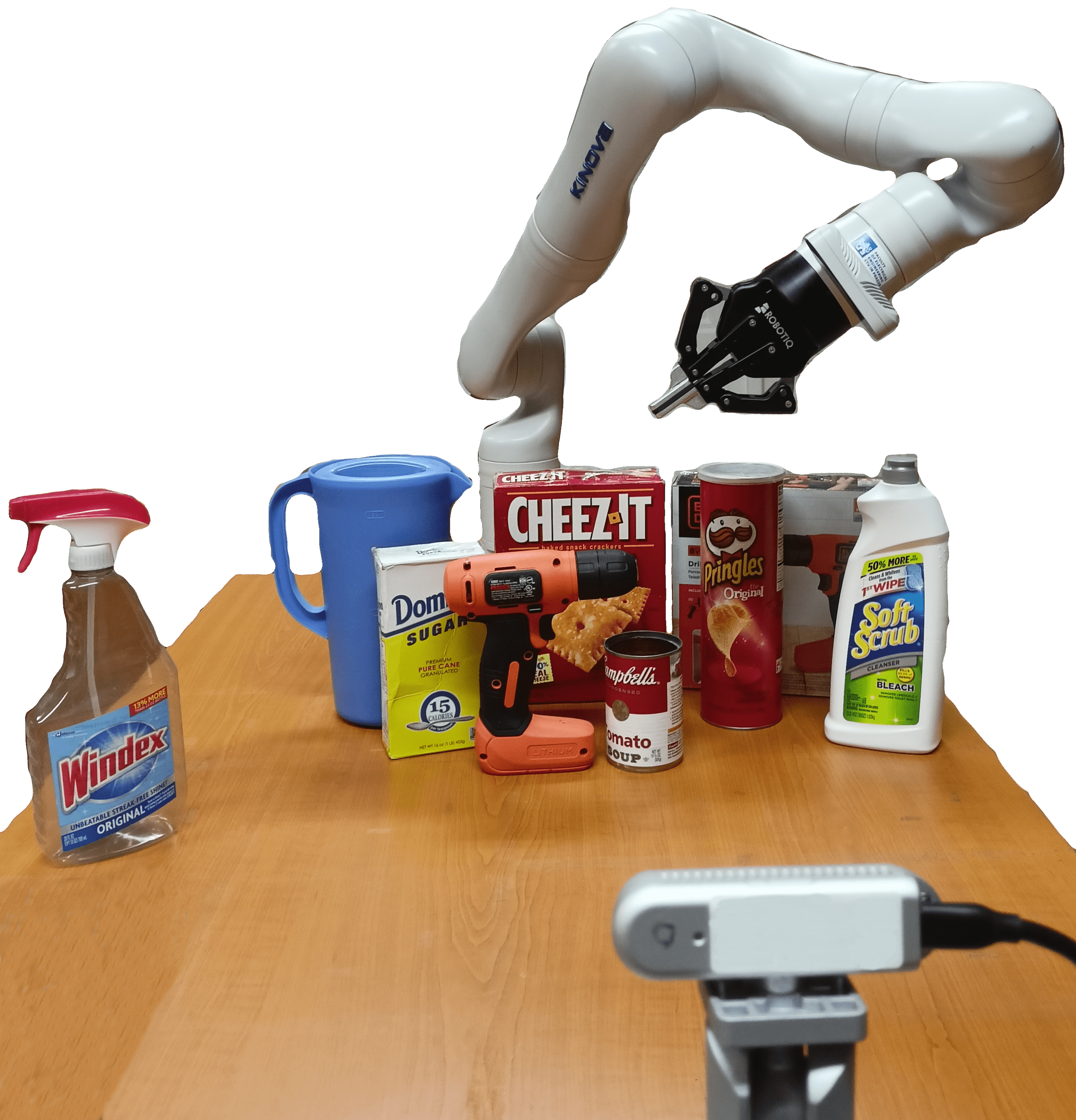}
    \caption{The real-world robot setup with Kinova Gen3 robot, Robotiq 2F-85 gripper, external RGB-D camera and all objects used. Closed gripper was used for haptic exploration, open for grasping.}
    \label{fig:setup}
    \customVspace{-1.5em}
\end{figure}

We evaluated the reconstruction obtained by \methodname{} on the eight objects shown in \figref{fig:setup} and on one more object for grasping---a transparent spray bottle for which \gt{} is not available, and thus it is not possible to compute the metrics.
Nevertheless, it is a typical object where haptic feedback dramatically improves the initial reconstruction from the RGB-D visual sensor. 
None of the objects were used in training the shape completion network.

In the real-world setup, we filled the objects with water to make them weigh about \SI{0.5}{kg} which simplifies collision detection by the robot. Note that in the experiments evaluating the Act-VH method~\cite{Rustler2022}, the objects were glued to the table and a dedicated finger was used. The \methodname{} method introduced a component that tracks object movement caused by touches (box (8) in Fig. 1). The sensitivity of the torque sensors is not sufficient for manipulation of very light objects. The current setup with objects partially filled with water is more challenging than gluing the objects. To avoid disadvantaging the reference method Act-VH, we report its results for objects glued to the table.

The arrangement of the experimental bench is shown in \figref{fig:setup}. The robot is a Kinova Gen3 with a Robotiq 2F-85 gripper and a Intel RealSense D435 camera. 

\subsection{Evaluation Metrics}
We used three metrics to evaluate accuracy: 
\begin{enumerate*}[label=(\roman*)]
    \item \ac{js}\ie{}the intersection over union of voxelized shapes; 
    \item \ac{cd}\ie{}the average distance of each point in one set to the closest point in the second set and vice versa;
    \item and the deviation of the reconstructed mesh surface area from the \gt{}.
\end{enumerate*}

We use three metrics, since the information they provide is complementary.
\ac{js} does not take into account the shape of the intersection and of the union\eg{}it attains the same value for a sharp hallucinated peak or a thin layer of added volume.
\ac{cd} is highly informative in most cases, but it is oversensitive to small scale changes, even if the reconstructed shape is close to \gt{}.
The deviation of the area of the mesh evaluates the accuracy of the estimated scale and allows to check for biases. 


Unless otherwise stated, experiments for each scene were repeated three times. We show the performance with real time on the x-axis, with individual touches numbered. 

\subsection{Simulation Experiments}
The simulation environment consists of a robot modeled in the MuJoCo~\cite{Todorov2012} simulator controlled through \ac{ros}. Objects are able to move on the table. Collision with the object is computed from the manipulator joint torques---as in the real setup.
Throughout this section, we compare the performance of \methodname{} with two variants of Act-VH~\cite{Rustler2022}: `Act-VH' and  `Act-VH -- new data'. Act-VH constitutes the original experimental results from \cite{Rustler2022}, in the setup with objects fixed to the table and poking with a dedicated probe, and \ac{igr} without any major changes. Only the results for the objects used in this paper were selected. This comparison also demonstrates improvements in the runtime of the methods.  `Act-VH -- new data' is a result of running the method from \cite{Rustler2022} on the data from the new setup---contacts with closed gripper and objects moving as a result of haptic exploration. This serves to isolate the benefits of the modifications of IGR inference in \methodname{}. Act-VH was run until 5 touches were completed; \methodname{} and `Act-VH -- new data' until 15 touches were done.

\subsubsection{Reconstruction -- one object in the scene}

\begin{figure}[h]
    \centering
    \begin{subfigure}[t]{0.49\textwidth}
        \centering
        \includegraphics[width=0.49\textwidth,trim={0.5cm 0 1.0cm 0},clip]{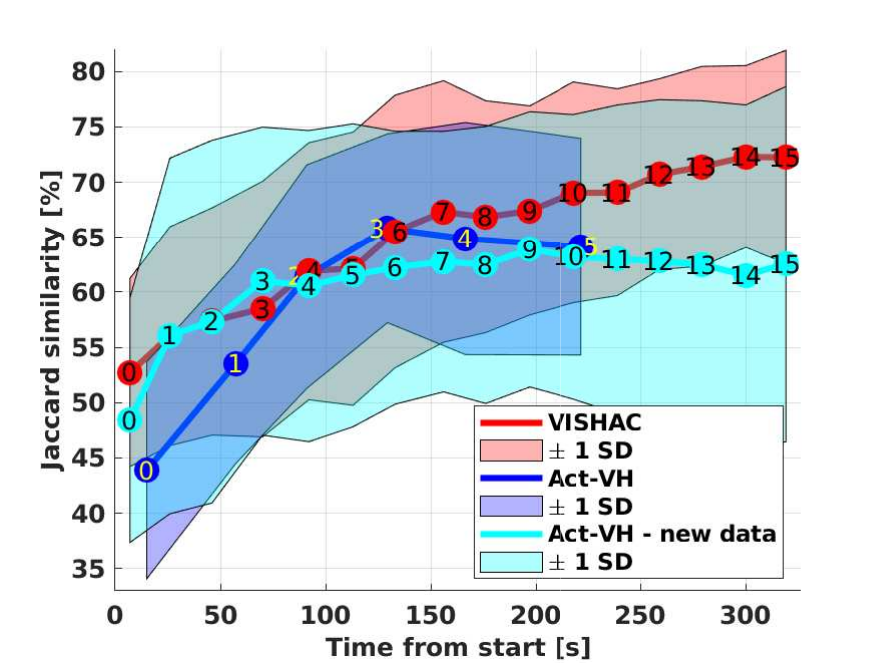}
        \includegraphics[width=0.49\textwidth,trim={0.5cm 0 1.0cm 0},clip]{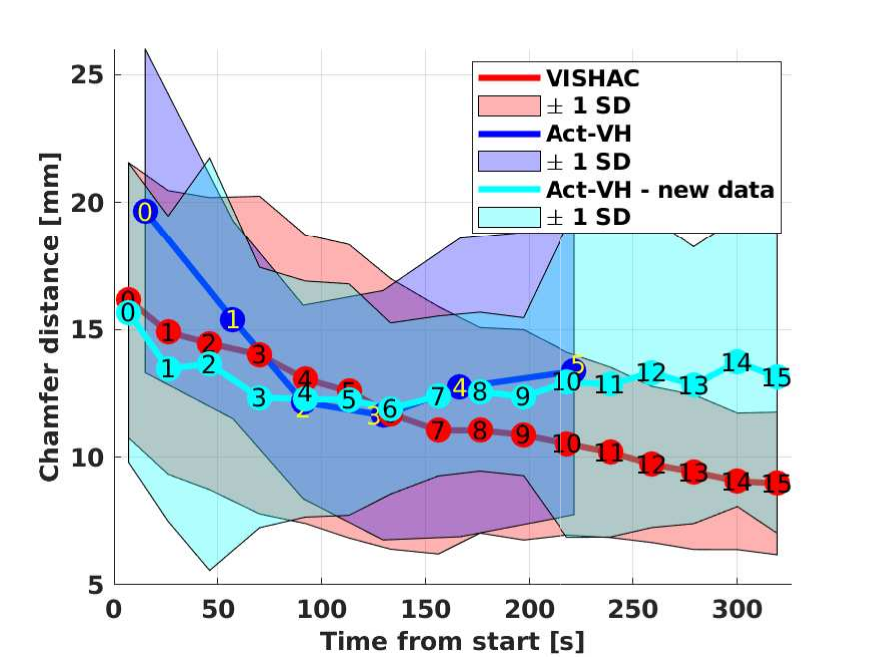}
    \end{subfigure}
    \caption{Simulation -- reconstruction -- 1 object in scene. Average reconstruction accuracy (8 objects, 3 repetitions each). Numbers in each datapoint -- number of touches. Shaded areas -- standard deviation. \acf{js} higher values better. \acf{cd} lower values better.}
    \label{fig:comparison_sim}
    \customVspace{-0.75em}
\end{figure}

A comparison of performance is shown in \figref{fig:comparison_sim}.
Both `Act-VH' and `Act-VH -- new data' are more ``greedy'' with higher reconstruction accuracy gains per touch. However, this comes at the expense of accuracy as more data come in. \methodname\newline is more ``conservative'', but maintains a steady performance gain. The relative increase in performance in the time when `Act-VH' completed 5 touches (approximately touch 10 in out method) is 7.7\% in \ac{js} and 21.3\% in \ac{cd}. After the last touch, \methodname{} is better than `Act-VH -- new data' with a relative difference of 15.5\% in \ac{js} and 31.7\% in \ac{cd}.

The same trend can be seen from the shaded areas, showing the confidence interval of $\pm 1$ standard deviation. The width of the areas shows the variability of the results for each object and each repetition of the experiment. For \methodname{}, the variability shrinks with time, showing that the method is more precise and robust. This is not the case for both versions of Act-VH.

\begin{figure}[tb]
    \centering
    \includegraphics[width=0.49\textwidth,trim={1.0cm 0 1.0cm 0},clip]{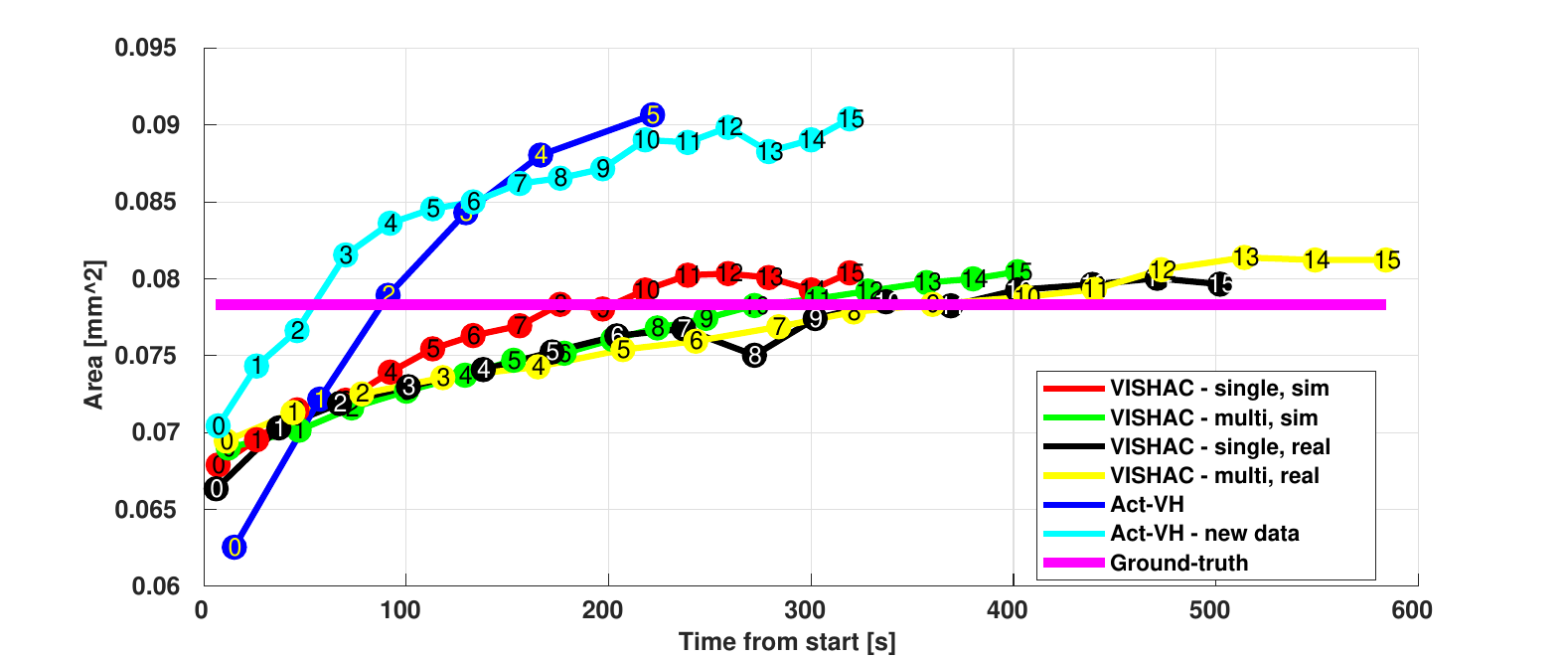}
    \caption{Simulation and real experiments. Mean area of meshes. Numbers in each datapoint -- number of touches. \textit{Single} -- scenes with only single objects;  \textit{multi} -- scenes with more objects. \textit{Act-VH} is a baseline from~\cite{Rustler2022} and \textit{Act-VH - new data} is the same method evaluated on data collected in this work.}
    \label{fig:area}
    \customVspace{-1.25em}
\end{figure}

In \figref{fig:area}, the deviations in the mesh area are shown. We can see that even though, for example, \ac{js} performance for Act-VH in simulation was similar to \methodname{} (see \figref{fig:comparison_sim}) there is a significant difference in area (blue for Act-VH, red for \methodname{}). The baseline method (evaluated on both new and original data) tends to inflate the shapes, resulting in good results for \ac{js} or \ac{cd}, but a high deviation in area. On the other hand, \methodname{} converges to the \gt{} value.

\subsubsection{Reconstruction -- multiple objects in the scene}

\begin{figure}[b]
    \centering
    \begin{subfigure}[t]{0.49\textwidth}
        \centering
        \includegraphics[width=0.49\textwidth,trim={0.5cm 0 0.0cm 0},clip]{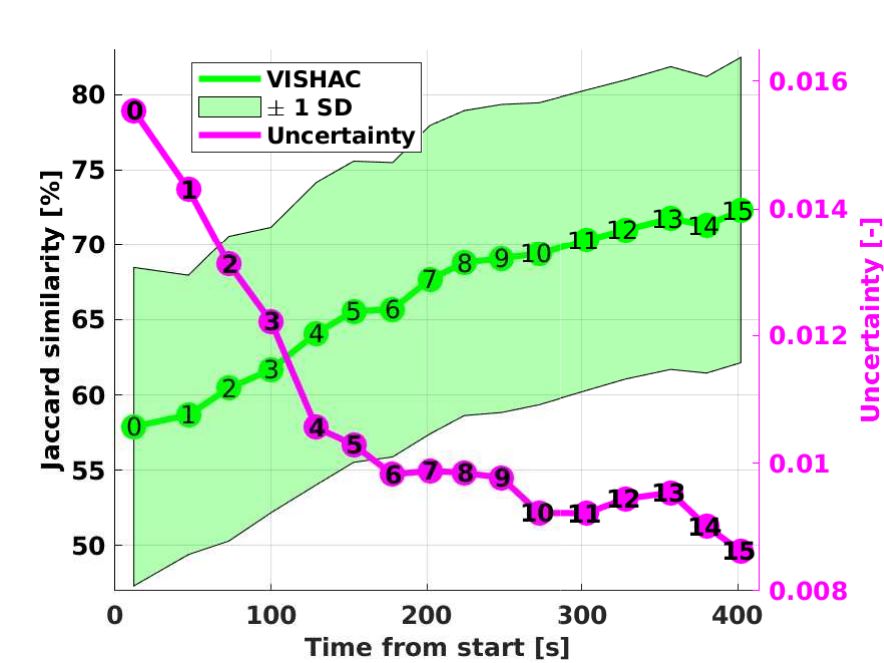}
        \includegraphics[width=0.49\textwidth,trim={0.5cm 0 0 0},clip]{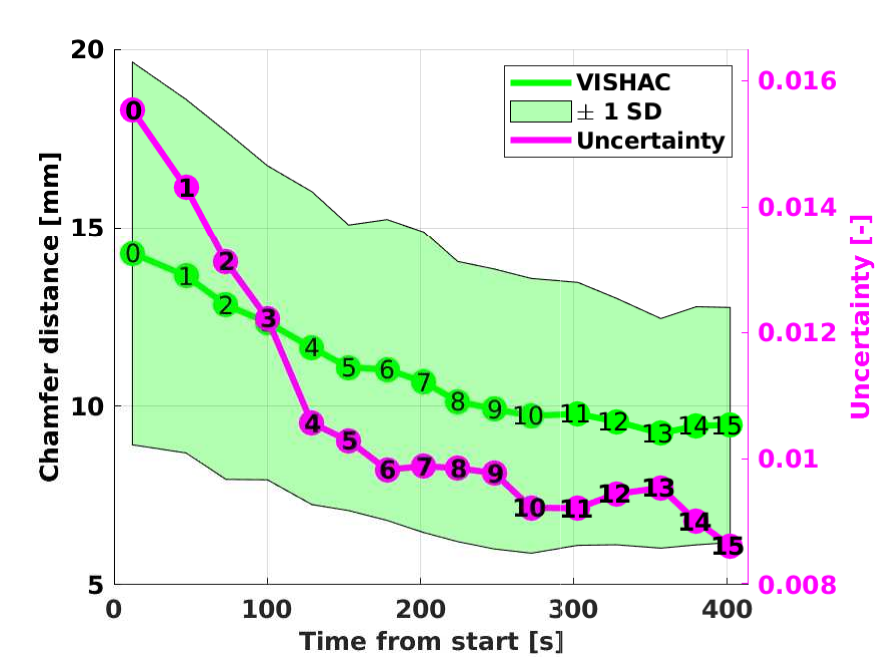}
    \end{subfigure}
    \caption{Simulation -- reconstruction -- multiple objects in scene. Average reconstruction accuracy (5 scenes, 3 repetitions each). Numbers in each datapoint -- number of touches. Shaded areas -- standard deviation. \acf{js} higher values better. \acf{cd} lower values better.}
    \label{fig:complex}
    \customVspace{-0.75em}
\end{figure}

In \figref{fig:complex} results for reconstructions of multiple objects in a more complex scenes are shown. We randomly selected five configurations of objects with two or three of them present in each scene.  
The accuracy is almost the same as that for one-object-at-a-time experiments. However, the runtimes for 15 touches are about \SI{80}{seconds} higher. This is mostly caused by more complicated touch point computation and motion planning. The deviations in the mesh area, shown in \figref{fig:area}, further prove that the pipeline behaves similarly with one or more objects. Overall, the results show that the pipeline is able to handle multiple objects at once. 

Furthermore, we show how the uncertainty (purple line) changes over time. The uncertainty is computed as the mean of the uncertainties for each point of each object. One can see that the uncertainty decreases with increasing accuracy. 
Thus, it could be used to evaluate the quality of reconstruction during runtime and stop the pipeline when a predefined criterion is met.


\subsection{Real-world Experiments}
We tested the pipeline on the same set of objects as in  simulation, in single- or multi-object configuration.
The superior performance of \methodname{} over Act-VH has already been demonstrated in simulation. Here we show the added value of \methodname{} in grasping experiments. 

\subsubsection{Reconstruction}

\begin{figure}[t]
    \centering
    \begin{subfigure}[t]{0.49\textwidth}
        \centering
        \includegraphics[width=0.49\textwidth,trim={0.5cm 0 1.0cm 0},clip]{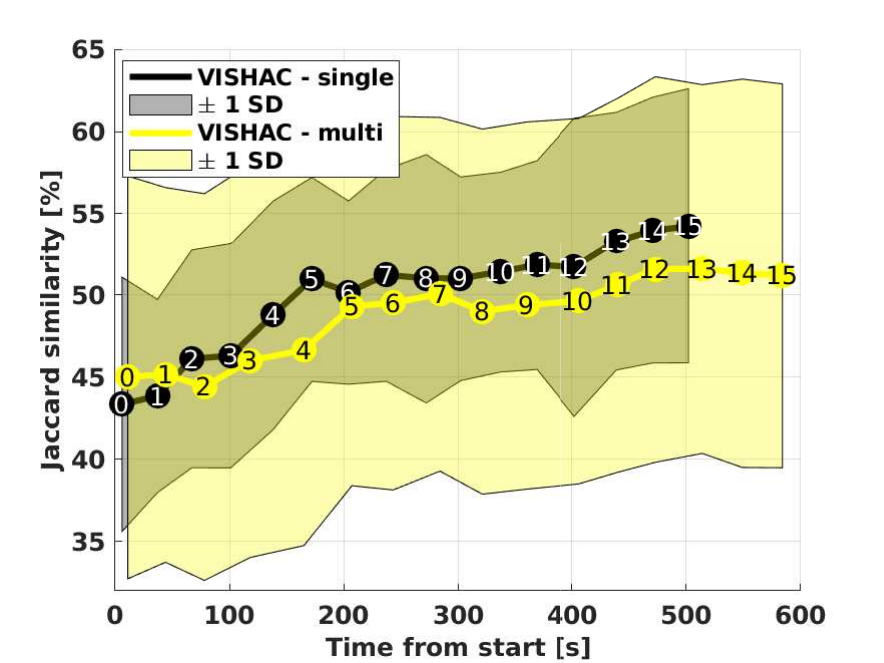}
        \includegraphics[width=0.49\textwidth,trim={0.5cm 0 1.0cm 0},clip]{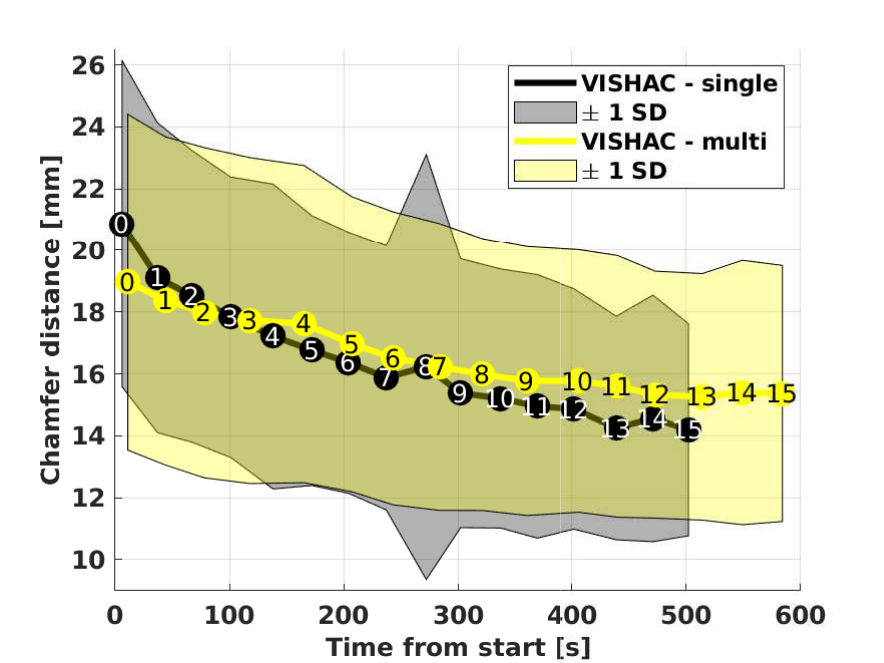}
    \end{subfigure}
    \caption{Real world -- Average reconstruction accuracy. `\methodname{} -- single' -- 8 objects, 3 repetitions each. `\methodname{} -- multi' -- 5 scenes, 3 repetitions each. Numbers in each datapoint -- number of touches. Shaded areas -- standard deviation. \acf{js} higher values better. \acf{cd} lower values better.
    }
    \label{fig:comparison_real}
    \customVspace{-1.5em}
\end{figure}

In \figref{fig:comparison_real}, the results for the precision of the reconstruction are shown. The single-object experiments are shown in black. We can see that the trends of both \ac{js} and \ac{cd} are the same as for the simulation, even though we can notice noise in some touches. In yellow, the results for multi-object experiments are shown. Again, the results for both types of experiments are similar, showing that the method is transferable to the real world. The overall accuracy in the real world is lower than in simulation. The main reason is noise in the RGB-D sensor and inaccurate collision detection (noise in the joint encoders).

The mean mesh area is shown in \figref{fig:area}. The area for single (black) and multiple (yellow) objects scenes approximately converges to the \gt{} value.

\subsubsection{Grasp Success Rate}

\begin{figure}[tb]
    \centering
    \includegraphics[width=0.45\textwidth,trim={0cm 0 0.0cm 0},clip]{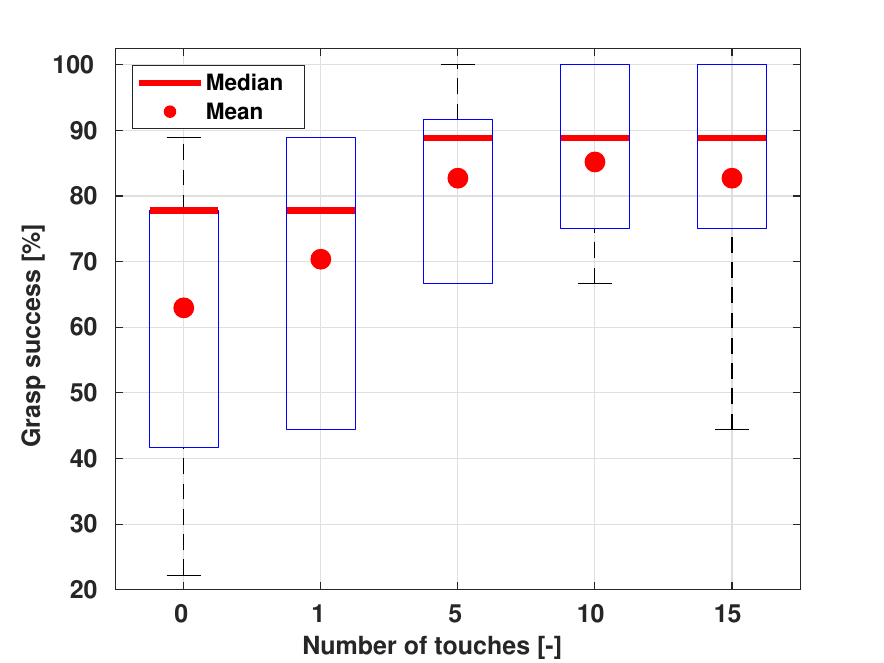}
    \caption{Real-world grasp success rate after different number of touches. Means over 3 grasp attempts on each of 3 repetitions on each object (9 grasps per object; 81 per box) are used to create the boxplots. Box edges indicates 25th and 75th percentiles, with whiskers showing extreme points. Medians (red line) and means (red circles) are computed over all 9 objects.}
    \label{fig:grasping}
    \customVspace{-1.75em}
\end{figure}

The last experiment evaluates the grasp success rate\ie{}the percentage of successful grasps. To sample grasp proposals, GraspIt!~\cite{Miller2004} was used. To check the quality of each grasp, the objects were picked and moved \SI{10}{cm} in the upwards direction. If the object did not fall from the gripper, the grasp was marked as successful. We decided to inspect grasp success using reconstruction after 0 and 1 touches to show how only a single touch improves the result. In addition, reconstructions after touches 5, 10, and 15 are used. We attempted to grasp 3 times for every repetition of the pipeline on each object, resulting in 9 grasp per object per touch---that makes 81 grasps per touch and 405 grasps in total. The results are shown in \figref{fig:grasping}. 

There is already a difference between 0 and 1 touches. The success rate increased from 63.3\% to 70.4\%. Maximum success was achieved using reconstructions after 10 touches. However, the difference between 5 and 15 touches is only 2.5\% (82.7\% vs. 85.2\%). 

In general, we can say that 5 touches are enough for a sufficient grasp success rate. To compare, the maximum success rate achieved in the baseline~\cite{Rustler2022} was 77.8\%. It is also worth mentioning that the result was achieved after time that could be comparable to touch number 12 in our results.

%% file: Sections/conclusion.tex
\section{Conclusion, Discussion, and Future Work}
We proposed a new method for shape completion using a combination of visual and haptic feedback.  \methodname{} outperformed the baseline Act-VH~\cite{Rustler2022} in terms of speed, reconstruction quality and robustness. 
We experimentally validated \methodname{} in both simulated and real-world environments, using 8 objects and an additional one for grasping.
\methodname{} was evaluated in scenes with one, two, or three objects. 
We always touched the objects 15 times and repeated each experiment three times, resulting in almost one hundred experiments in total.
In addition, a new uncertainty computation strategy was evaluated, showing that it can be used for on-the-fly quality measurements. The reconstructions were furthermore validated with more than 400 grasps, demonstrating the usability of shape completion in a core robotic task.

There are several directions for future work. 
The results in the real setup are negatively affected by the noise induced by the contact events---the collision is detected with a certain delay, the object moves, and the new pose is not re-estimated perfectly. This could be mitigated in two ways. First, the most effective will be faster contact detection. In the current setup, collisions are detected from the joint torque sensors in the manipulator and its dynamic model and their remapping onto the end effector. In our setup, this leads to delayed and noisy estimation of the collision and significant movement of the object. A remedy would be a force/torque sensor at the robot wrist or tactile sensors at the end effector. 
Second, the object pose re-estimation after every haptic exploration could be further improved by using alternative pose estimators or adding tracking. 

Furthermore, on a robot hand with sensorized fingertips, data for reconstruction could be collected more effectively by sliding the fingers over the object surface (tactile servoing). Finally, poking and touching reveal surface stiffness and other physical properties that play a role in grasping and could be exploited. 
